\documentclass[]{article}
\usepackage{proceed2e}
\usepackage{natbib}
\usepackage{amsmath,amssymb}
\usepackage{amsthm,amsmath}
\usepackage{xcolor,colortbl}

\newcommand{\ci}[3]{\ensuremath{I({#1},{#2} \mid #3})}

\newcommand{\triplet}[3]{\ensuremath{\langle{#1},{#2}\mid #3}\rangle}

\newcommand{\I}{\ensuremath{\mathcal{I}}}
\newcommand{\F}{\ensuremath{\mathcal{F}}}

\newcommand{\project}[2]{#1\langle#2\rangle}
\newcommand{\f}[2]{\delta(\project{x}{#1},#2)}

\newtheorem{auxiliarylemma}{Auxiliary Lemma}
\newtheorem{theorem}{Theorem}

\newtheorem{corollary}{Corollary}
\newtheorem{definition}{Definition}
\newtheorem{proposition}{Proposition}

\DeclareMathOperator{\val}{Val}

\title{Markov random fields factorization with context-specific
  independences}

\author{ {\bf Alejandro Edera} \\  
Dep. de Sistemas de Computaci\'on \\  
Universidad Tecnol\'ogica Nacional\\ 
\And 
{\bf Facundo Bromberg}  \\ 
Dep. de Sistemas de Computaci\'on \\  
Universidad Tecnol\'ogica Nacional\\ 
\And 
{\bf Federico Schl\"uter}   \\ 
Dep. de Sistemas de Computaci\'on \\  
Universidad Tecnol\'ogica Nacional\\ 
} 

\begin{document}

\maketitle 

\begin{abstract}
Markov random fields provide a compact representation of joint probability distributions
by representing its independence properties in an undirected graph.
The well-known Hammersley-Clifford theorem uses these conditional 
independences to factorize a Gibbs distribution into a set of factors.
However, an important issue of using a graph to represent independences is that
it cannot encode some types of independence relations, such as the context-specific independences (CSIs).
They are a particular case of conditional independences that 
is true only  for a certain assignment of its conditioning set; in 
contrast to conditional independences that must hold for all its assignments.
This work presents a method for factorizing a Markov random field
according to CSIs present in a distribution, and formally guarantees that this factorization is correct.
This is presented in our main contribution, the context-specific Hammersley-Clifford theorem,
a generalization to CSIs of the Hammersley-Clifford theorem that applies for conditional independences.
\end{abstract}

\section{Introduction}
\emph{Markov random fields} (MRFs), also known as undirected graphical
models, or Markov networks, belong to the family of probabilistic
graphical models \citep{koller09}, a well-known computational
framework for compact representation of joint probability
distributions.  These models are composed of an independence
structure, and a set of numerical parameters.  The independence
structure is an undirected graph that encodes compactly the
conditional independences among the variables in the domain. Given the
structure, the numerical parameters quantify the relationships in the
structure.
Probability distributions present in practice important complexity deficiencies, with
exponential space complexity of their representation, time complexity of inference,
and sample complexity when learning them from data.
Based on the structure of independences, it is possible to represent efficiently
the joint probability distribution by factorizing it into smaller functions (or factors),
each over a subset of the domain variables,
resulting some times in exponential reductions in these complexities.
This factorization can be done by using the
well-known Hammersley-Clifford theorem
\citep{Hammersley_Clifford_1971}.
%

An important issue of using a graph to represent independences is that
it cannot encode some types of independence relations, such as the
\emph{context-specific independences} (CSIs)
\citep{DBLP:conf/uai/BoutilierFGK96}.
These independences are similiar to conditional independences except
that are only true for certain assignments of its conditioning set.
The CSIs have been applied in a
wide range of scenarios achieving significant improvements in time,
space and sample complexities, in comparison with other approaches
that only uses conditional independences encoded by the graph.
\citep{chikering-abatlbnwls,fierens-csiidrpmaiioteogs,
  DBLP:journals/jair/PooleZ03,wexler-ifmm,lowd2010learning,ravikumar2010:l1}.
In these contributions, the CSIs are encoded in alternative data structures
(e.g., using a decision tree instead of a graph). This is carried out
by assuming that the factors of the distribution are conditional
probability distributions. In this sense, the CSIs are not used to
factorize the distribution, but they are used for representing
efficiently the factors.

The main contribution of our work is the \emph{context-specific Hammersley-Clifford} theorem. 
The importance of this theoretical result lies in that it allows to
factorize a distribution using CSIs, to obtain a more sparse representation than
that obtained with conditional independences, providing theoretical guarantees.
For this, a log-linear model is used as a more fine-grained representation of the MRFs \citep{koller09}.
By using such models it is possible to extend the
advantages of the Hammersley-Clifford theorem, that is, improvements
in time, space and sample complexities. 

The remainder of this work is organized as follows.
The next Section provides a summary of the related work in the literature.
Section~\ref{sec:preliminaries} presents an overview of how to
factorize a distribution by exploiting its independences. 
Section~\ref{sec:llmodels} formally describes the context-specific
Hammersley-Clifford theorem that factorizes a log-linear model
according to a set of CSIs.
The paper concludes with a summary in Section~\ref{sec:conclusions}.

\section{Related work}

There are several works in the literature \citep{PietraPL97,Lee+al:NIPS06,lowd2010learning,vanhaaren2012} that 
learn log-linear models directly by presenting different procedures for selecting features from data. 
Neither of these works discuss CSIs, nor present any guarantee on how the
log-linear model generated is related to the underlying distribution.

CSIs were first introduced by \citep{DBLP:conf/uai/BoutilierFGK96}
by coding them locally within conditional probability tables (factors) of Bayesian networks
as decision trees. 
Their approach is hybrid, encoding conditional independencies in the directed graph and
CSIs as decision trees over the variables of a 
conditional probability table. Also, their work
presents theoretical results for a sound graphical representation.
This work instead proposes a unified representation
for CSIs and conditional independencies into a log-linear model. As such, it requires
first theoretical guarantees on how a distribution factorizes according to this model
(not needed for the work of Boutlier as the factorization into conditional probability
tables is not affected by the CSIs). 
It remains for future investigation to find an efficient graphical representation 
(and theoretical guarantees thereon). 

The work of \citep{gogate2010learning} is the closests to our work, presenting
an algorithm for factorizing a log-linear model according to CSIs. For that it 
introduces a statistical independence test for eliciting this independencies from
data. The work assumes the underlying distribution to be a \emph{thin junction tree}.
Although some theoretical results are presented that guarantee an efficient
computational performance, no results are presented that guarantee the factorization
proposed is sound.

\section{Preliminaries}\label{sec:preliminaries}

%
This section provides some background on MRFs, 
explaining how to factorize a distribution by exploiting its independences. 
Let us start by introducing some necessary notation.
We use capital letters for sets of indexes, reserving the $X$ letter for the domain of a distribution,
and \(V\) for the nodes of a graph.
Let $X = (X_a, X_b, \ldots, X_n)$ represent a vector of $n=|X|$ random variables.
The $\val(X_a)$ function returns all the values of the domain of $X_a$, 
and $\val(X_U)$ returns all the possible values of the set of variables \(X_U = (X_i, i \in U)\).
Let \(x = (x_a, x_b, \ldots, x_n)\) be a complete assignment of \(X\).
The values of \(X_a\) are denoted by \(x_a^j\in\val(X_a)\),
where \(j=1,\ldots,|\val(X_a)|\).
Finally, we denote by \(\project{x}{W}\) the value taken by variables \(X_W\) in the complete assignment \(x\).

Conditional independences are regularities of
distributions that has been extensively studied in the field of
statistics, demonstrating how they can be effectively and soundly used
for reducing the dimensionality of the distribution \citep{pearl88,Spirtes00,koller09}.
Formally, a conditional independence is defined as follows:

\begin{definition}{\textbf{Conditional independence}}.
  Let $X_a , X_b \in X$ be two random variables, and $X_U \subseteq X \setminus \{X_a,X_b\}$ be a set of variables.
  We say that \(X_a\) and \(X_b\) are \emph{conditionally independent} given \(X_U\),
  denoted as $\ci{X_a}{X_b}{X_U}$,
  if and only if for all values \(x_a\in\val(X_a)\), \(x_b\in\val(X_b)\), and \(x_U\in\val(X_U)\):
  \begin{align} \label{eq:ci}
    p(X_a \vert X_b, X_U) = p(X_a \vert X_U),
  \end{align}
  whenever \(p(X_b, X_U) > 0\).
\end{definition}

Through the notion of conditional independence it is possible to
construct a \emph{dependency model} \I, defined formally as follows:
\begin{definition}{\textbf{Dependency model}}.\label{def:dm}

A dependency model $\I$ is a discrete function that returns a truth value,
given an input triplet $\triplet{X_a}{X_b}{X_U}$, for all
$X_a , X_b \in X$, $X_U \subseteq X \setminus\{X_a,X_b\}$. 
\end{definition}

\paragraph{Remark.} 
An alternative viewpoint of the above definition can be obtained by considering 
that every triplet $\triplet{X_a}{X_b}{X_U}$ over a domain $X$ are implicitly
conditioned by a constant assignment to some external variable of the domain \(E=e\).
In this sense, all the triplets of the dependency model become to be conditioned by the assignment \(E=e\).

In that sense, any probability distribution is a dependency
model, because for any conditional independence assertion it is
possible to test its truth value using Equation~\eqref{eq:ci}.  In this
work, we are particularly interested in the set of dependency models
that are \emph{graph-isomorph}, that is when all its independences and dependences
can be represented in an undirected graph.
Formally, an undirected
graph \(G = (V,E)\) is defined by a set of nodes \(V = (a,b,\ldots,
n)\), and a set of edges \(E \subset V\times V\). Each node \(a\in V\)
is associated with a random variable \(X_a\in X\), and each edge
\((a,b)\in E\) represents a direct probabilistic influence between \(X_a\) and \(X_b\).
A necessary and sufficient condition for dependency models to be graph-isomorph is that
all its independence assertions satisfy the following independence axioms,
commonly called the \emph{Pearl axioms} \citep{PearlPaz1985:graphoids}:

\vspace{2ex}
\scriptsize
\noindent
Symmetry
\begin{equation}
\ci{X_A}{X_B}{X_U} \Leftrightarrow \ci{X_B}{X_A}{X_U}
\label{symmetry}
\end{equation}
Decomposition
\begin{equation}
\begin{split}
\ci{X_A}{X_B~\cup~X_W}{X_U} \Rightarrow \ci{X_A}{X_B}{X_U} ~\&~\ci{X_A}{X_W}{X_U}
\label{decomposition}
\end{split}
\end{equation}
Intersection
\begin{equation}
\begin{split}
\ci{X_A}{X_B}{X_U~\cup~X_W}~\&~ \ci{X_A}{X_W}{X_U~\cup~X_B} \Rightarrow \\ \ci{X_A}{X_B~\cup~X_W}{X_U}
\label{intersection}
\end{split}
\end{equation}
Strong union
\begin{equation}
\ci{X_A}{X_B}{X_W} \Rightarrow \ci{X_A}{X_B}{X_W\cup X_U}
\label{weak}
\end{equation}
Transitivity
\begin{equation}
\ci{X_A}{X_B}{X_W} \Rightarrow \ci{X_A}{X_c}{X_W} ~\mbox{or}~ \ci{X_c}{X_B}{X_W}
\label{contraction}
\end{equation}

\normalsize

Other important property that we will need later to reconstruct graphs 
from dependency models is the pairwise Markov property, that asserts that
an undirected graph can be built from a dependency model which is graph-isomorph, as follows:

\begin{definition}[Pairwise Markov property \citep{koller09}]
  \label{def:pwp}
  Let $G$ be a graph over \(X\). Two nodes $a$ and $b$ are
  non-adjacent if and only if the random variables $X_a$ and $X_b$ are
  conditionally independent given all other variables $X \setminus
  \{X_a,X_b\}$, i.e., 

  \begin{equation}
    \ci{X_a}{X_b}{X \setminus\{X_a,X_b\}} \text{ iff }	(a,b)\notin E.
  \end{equation}
\end{definition}

If every independence assertion contained in a dependency model \(\I\) holds for \(p(X)\),
\(\I\) is said to be an \emph{I-map} of \(p(X)\).
In a similar fashion, we say that $G$ is also an I-map of \(p(X)\).
The pairwise property is necessary for those cases for which the graph can only encode a subset of
the independences present in the distribution.

A distribution can present additional type of independences. In this work we 
focus in a finer-grained type of independences: the \emph{context-specific independences} (CSI)
\citep{DBLP:conf/uai/BoutilierFGK96,geiger-heckerman-kraiisnabm,chikering-abatlbnwls,koller09}.
These independences are similar to conditional independences,
but hold for a specific assignment of the conditioning set, called
the \emph{context} of the independence.
We define CSIs formally as follows:

\begin{definition}[Context-specific independence
  \citep{DBLP:conf/uai/BoutilierFGK96}]

  Let $X_a , X_b \in X$ be two random variables, $X_U,X_W \subseteq X \setminus\{X_a,X_b\}$
  be pairwise disjoint sets of variables that does not contain $X_a,X_b$;
  and $x_W$ some assignment of $X_W$.
  We say that variables \(X_a\) and \(X_b\) are
  \emph{contextually independent} given \(X_U\) and
  a context $X_W=x_W$, denoted $\ci{X_a}{X_b}{X_U,x_W}$, if and only if
  \begin{align} \label{eq:csi}
    p(X_a \vert  X_b, X_U, x_W) = p(X_a \vert X_U, x_W),
  \end{align}
  whenever \(p(X_b, X_U, x_W) > 0\).
\end{definition}

Interestingly, a conditional independence assertion can be seen 
as a conjunction of CSIs, that is, the CSIs for all the contexts of the conditioning set of
the conditional independence.
Since each CSIs is defined for a specific context, 
they cannot be represented all together in a single undirected graph \citep{koller09}.
Instead, they can be captured by a dependency model $\I$, extended for CSIs
by using Equation~\eqref{eq:csi} to test the
validity of every assertion $\ci{X_a}{X_b}{X_U, x_W}$.
We call this model a \emph{context-specific dependency model} \(\I_c\).
If every independence assertion contained in \(\I_c\) holds for \(p(X)\),
\(\I_c\) is said to be an \emph{CSI-map} of \(p(X)\) \citep{DBLP:conf/uai/BoutilierFGK96}.
We define formally the Context-specific dependency model as follows:
\begin{definition}{Context-specific dependency model}.\label{def:csdm}
A dependency model $\I_c$ is a discrete function that returns a truth value
given an input triplet $\triplet{X_a}{X_b}{X_U,x_W}$,
for all $X_a,X_b \in X$, $X_U \subseteq X \setminus\{X_a,X_b\}$, 
and \(x_W\) a context over the subset \(X_W\subseteq X\).
\end{definition}

\subsection{Undirected graphs factorization}\label{sec:markov-networks}
A MRF uses an undirected graph $G$ and a set of numerical parameters
$\theta\in\mathbb{R}$ to represent a distribution. The completely connected sub-graphs 
of \(G\) (a.k.a., \emph{cliques}) can
be used to factorize the distribution into a set of \emph{potential
  functions} \(\{\phi_C(X_C)~ \colon~ C\in cliques(G))\}\) of lower dimension than
\(p(X)\), parameterized by \(\theta\). The following
theorem shows how to factorize the distribution:

\begin{theorem}[Hammersley-Clifford \citep{Hammersley_Clifford_1971}] \label{thm:HC}
  Let $p(X)$ be a positive distribution over the domain of variables $X$,
  and let $G$ be an undirected graph over $X$. If $G$ is an I-map of
  $p(X)$, then $p(X)$ can be factorized as:
  \begin{align} \label{eq:Gibbs}
    p(X) &= \exp\{\sum_{C \in cliques(G)} \phi_C(X_C) - \ln(Z)\},
  \end{align}
  where \(Z\) is a normalizing constant.
\end{theorem}

A distribution factorized by the above theorem is called a \emph{Gibbs
  distribution}. The most n\"aive form contains potentials
\(\phi_C(\cdot)\) represented by tables, where each entry
corresponds to an assignment \(x_C\in\val(X_C)\) that has associated a numerical parameter.

Despite the clear benefit of the factorization described by the
Hammersley Clifford theorem, the representation of a factor as a
potential does not allow to encode CSIs. These patterns are more
easily encoded in a more convenient representation called
\emph{log-linear}. The log-linear model represents a Gibbs
distribution by using a set of \emph{features} \F\ to represent the
potentials. A feature is an assignment to a subset of variables of
domain. We denote a features as \(f_C^j\), to make more clear the
distinction between the features of a log-linear and its input
assignment \(x\). Thus, a potential in a log-linear is represented as
a linear combination of features as follows:

\[
\phi_C(X_C = \project{x}{C}) = \sum_j^{|\val(X_C)|} \theta_{j}\f{C}{f_C^j},
\]

where $\f{C}{f_C^j}$ is the Kronecker delta function, that is, it
equals to $1$ when $\project{x}{C} = f_C^j$, and $0$ otherwise. By
joining the linear combinations of all the potentials and merging its
indexes into a unique index $\alpha\in\{1,\ldots,|\F|\}$,
we can represent Equation~\eqref{eq:Gibbs} by using the following
log-linear model:

\begin{align} \label{eq:log-linear-indicator}
  p(X=x) &= \exp\{\sum_\alpha \theta_{\alpha}\f{C
_\alpha}{f_{C_\alpha}^\alpha} - \ln(Z)\}.
\end{align}

In the next section we present the context-specific
Hammersley-Clifford theorem, a generalization of the Hammersley Clifford theorem
that shows how to factorize a distribution (represented by a
log-linear) using a context-specific dependency model \(\I_c\) that
captures the CSIs. 

\section{Context-specific Hammersley-Clifford} \label{sec:llmodels}

This section presents the main contribution of this work: a
generalization of the Hammersley-Clifford theorem 
for factorizing a distribution
represented by a log-linear based on a context-specific dependency model
\(I_c\) CSI-map of \(p(X)\). For this, we begin by defining the following
Corollary of the Hammersley-Clifford theorem:

\begin{corollary}[Independence-based
  Hammersley-Clifford] \label{thm:HC2}

  Let $p(X)$ be a positive distribution, and let $\I$ be a 
  graph-isomorph dependency model over \(X\). If \I\ is an I-map of $p(X)$, then
  \(p(X)\) can be factorized into a set of potential functions
  $\{\phi_{C_i}(X_{C_i})\}_i$, such that for any $\ci{X_a}{X_b}{X_W}$
  that is true in \I, there is no factor $\phi_i(X_{C_i})$ that
  contains both variables \(X_a\) and \(X_b\) in \(X_{C_i}\).

  \begin{proof}
    From the assmuptions, \I\ is graph-isomorph and is an I-map of $p(x)$.
    By definition, the former implies there exists an
    undirected graph \(G(V,E)\) that exactly encodes \I, and it 
    therefore must also be I-map of $p(x)$. The assumptions of 
    the Hammersley-Clifford Theorem~\ref{thm:HC} hold, so \(p(X)\)
    can be factorized into a set of potential functions over the cliques of \(G\). 
    Also, since
    \I\ is graph-isomorph, its conditional
    independences satisfy the Pearl axiom, in particular the strong union
    axiom. Therefore if conditional independence $\ci{X_a}{X_b}{X_W}$
    is in \I, the conditional independence
    $\ci{X_a}{X_b}{X\setminus{X_a,X_b}}$ is also in \I. Using this fact in the
    pairwise Markov property we can imply the
    no-edge $(a,b)\notin E$; in other words, $a$ and $b$ cannot belong
    to the same clique. Since Hammersley-Clifford holds, this last fact
    implies no factor $\phi_i(X_{C_i})$ can
    contain both variables \(X_a\) and \(X_b\) in \(X_{C_i}\).
  \end{proof}
\end{corollary}

This corollary shows how to use a dependency model \I\ (instead of a graph) 
to factorize the
distribution \(p(X)\). In what follows, we present theoretical results
that show how a context-specific dependency model \(\I_c\) can be used to factorize
$P(X)$. The general rationale is to decompose \(\I_c\) into subsets of CSIs contextualized
on certain context $x_W$ that are
themselves dependency models over sub-domains, 
and use those to decompose the conditional distributions of $p(X)$ using Hammersley-Clifford. 

\begin{definition}[Reduced dependency model]
  Let \(p(X)\) be a distribution over \(X\), \(x_W\) a context
  over subset \(X_W\subseteq X\), and \(\I_c\) a
  context-specific dependency model over $X$. 
  We define the \emph{reduced dependency model} \(\I_{x_W}\) of \(\I_c\) over domain $X\setminus X_W$ as the rule that 
  for each $X_a, X_b \in X$, each pair $X_U,X_W$ of disjoint subsets of $X\setminus\{X_a,X_b\}$,
  and each assignments $x_W$ of $X_W$, assigns 
  a truth value to a triplet $\triplet{X_a}{X_b}{X_U,x_W}$ from independence assertions in \(\I_c\) as follows:

  \begin{align} \label{eq:conjunction}
 & \I_{x_W} ( \triplet{X_a}{X_b}{X_U,x_W} ) =  \\ &  \bigwedge_{X_U \in Val(X_U)} \I_c( \triplet{X_a}{X_b}{x_U,x_W} ) \nonumber
  \end{align}
\end{definition}


The following proposition relates the CSI-mapness of a context-specific
dependency model and the I-mapness of its reduced dependency models.

\begin{proposition}\label{prop:1}
  Let \(p(X)\) be a distribution over \(X\), \(x_W\) be a context
  over subset of \(X\), and \(\I_c\) be a context-specific dependency
  model over \(X\). If \(I_c\) is a CSI-map of \(p(X)\), then
  \(\I_{x_W}\) is an I-map of \(p(X\setminus X_W \mid x_W)\).

  \begin{proof}
    We start arguing that \(\I_{x_W}\)  is a CSI-map of $p(X)$, and then extend
    the proof to show that it is an I-map of the conditional \(p(X \setminus X_W \mid x_W)\).
    That \(\I_{x_W}\)  is a CSI-map of $p(X)$ follows from the fact that \(\I_c\)
    is a CSI-map of $p(X)$, that implies that not only its CSIs holds in $p(X)$, but
    any CSI obtained by conjoining those CSIs over all values of any of its variables, in particular
    the conjunction of Equation~\eqref{eq:conjunction}. 
    That \(\I_{x_W}\)  is an I-map of \(p(X \setminus X_W\mid x_W)\) follows from the fact that any 
    CSI $I(X_a,X_b \mid X_U,x_W)$ in $p(X)$ is equivalent to a conditional independence $I(X_a,X_b \mid X_U)$
    in the conditional \(p(X\setminus X_W\mid x_W)\).
%
  \end{proof}
\end{proposition}

In the next auxiliary lemma it is shown how to factorize a distribution \(p(X)\)
using a dependency model \(\I_{x_W}\):

\begin{auxiliarylemma}
  \label{lem:1}
  Let $p(X)$ be a positive distribution over \(X\), $I_c$ be a dependency model over $X$,  
  and \(\I_{x_W}\) be a graph-isomorph dependency model
  over $X \setminus X_W$. If \(\I_{x_W}\) is an I-map of the conditional 
 \(p(X \setminus X_W \mid x_W)\), then this conditional can be factorized into a set
  of potential functions \(\{\phi_i(X_{C_i})\}_i\) over $X \setminus X_W$, 
  such that for any $\ci{X_a}{X_b}{X_U, x_W}$ that is true in
  \(\I_{x_W}\), there is no factor \(\phi_i(X_{C_i})\) that contains both
  \(a\) and \(b\) in \(C_i\).


  \begin{proof}
    The proof consists on using Corollary~\ref{thm:HC2} for the conditional 
  \(p(X\setminus X_W \mid x_W)\) as the distribution, and $I_{x_W}$ as 
  the dependency model. For that, we show they satisfy the requirements of the Corollary, 
  that is, \(p(X\setminus X_W \mid x_W)\) is positive, and $I_{x_W}$ is a graph-isomorph
  dependency model over domain $X \ X_W$ that is an I-map of the conditional.  
  The $I_{x_W}$ is an I-map of the conditional and  
  graph-isomorph follows from the assumptions. It remains to prove then 
  the positivity of the conditional. For that, the conditional is expanded as follows:

    \begin{align*}
      & p(X \setminus \{X_W\} \mid x_W) = \frac{p(X \setminus \{X_W\}, x_W)}{p(x_W)} \\
      &= \frac{p(X \setminus \{X_W\}, x_W)}{\sum_{x_{X\setminus X_W}\in\val(X \setminus \{X_W\})} p(x_{X\setminus W},x_W)},&
    \end{align*}

    where the sum expansion of the denominator follows from the law of
    total probability.  The conditional has been expressed then as an
    operation over joints, and being all positive, it follows that
    both the numerator and denominator, and therefore the whole
    quotient is positive.
  \end{proof}
\end{auxiliarylemma}


With Lemma \ref{lem:1}, we can present our main theoretical
result, a theorem that generalizes Theorem~\ref{thm:HC} to factorize
the features $\F$ in a log-linear of \(p(X)\) according to some given context-specific dependency model  \(\I_c\).
For this, we need to define precisely what we mean by 
factorization of a set of features \F. We do this in two steps,
one that defines a factorization according to dependency models,
and then the contextualized case for context-specific dependency models.

\begin{definition}[Feature factorization]
  \label{def:dmf}
  Let \F\ be a set of features over some domain $X$, and \(\I_{x_W}\) 
  some reduced dependency model over $X \ X_W$. We say 
  features \F\ \emph{factorize} according to \(\I_{x_W}\) if
  for each  $\ci{X_a}{X_b}{X_U, x_W}$ that is true in \(\I_{x_W}\), and each feature
  $f_C \in \F$ such that   $\project{f_C}{W} = x_W$, it holds that
  either \(a\notin C\) or  \(b\notin C\).
\end{definition}


\begin{definition}[Context-specific feature factorization]
  \label{def:ff}

  Let \F\ be a set of features over some domain $X$, and \(\I_c\) be a
  context-specific dependency model.
  The features \F\ are said to \emph{factorize} according to \(\I_c\) if
  they factorize according to each reduced dependency model $\I_{x_w}$ of \(\I_c\)
  (as defined by Definition~\ref{def:dmf}),
  with $X_W \subseteq X$, and  $x_W \in \val(X_W)$.


\end{definition}

We present now our main theorem, and then discuss practical issues regarding its requirements.

\begin{theorem}[Context-Specific Hammersley-Clifford]
  \label{thm:CSHC}
  Let $p(X)$ be a positive distribution over \(X\), \F\ be a set of
  features from a log-linear of \(p(X)\), and  \(\I_c\) be a context-specific dependency model over $X$,
  such that each of its reduced dependency models (over all possible contexts) is
  graph-isomorph.  If  \(\I_c\) is CSI-map of \(p(X)\) then \F\ factorizes according
  to  \(\I_c\).
\begin{proof}
  From the definition of context-specific feature factorization, the conclusion of
the theorem holds if $\F$ factorizes according to each reduced dependency model of $\I_c$. 
So let $\I_{x_W}$ be some arbitrary reduced dependency model for context $x_W$, and 
prove $\F$ factorizes according to $\I_{x_W}$, which by Definition~\ref{def:dmf} requires that
(a) for each $\ci{X_a}{X_b}{X_u,x_W}$ that is true in $\I_{x_W}$, 
and (b) for each $f_C \in \F$ s.t. $\project{f_C}{W} = x_W$, it holds that (c) either 
$a \notin C$ or $b \notin C$.

To proceed then, we first apply the Auxiliary Lemma~\ref{lem:1} for $p(X)$, the context $x_W$,
and the reduced dependency model $\I_{x_W}$. 
These requirements are satisfied, that is, 
$p(X)$ is positive and $\I_{x_W}$ is both graph-isomorph and I-map of the conditional 
\(p(X\setminus X_W \mid x_W)\) (by Proposition~\ref{prop:1}). From this we conclude
the consequent of the Lemma, i.e., that the conditional
$p(X \setminus X_W \mid x_W)$ can be factorized into a set of potencial functions \(\{\phi_i(X_{C_i})\}_i\)  s.t. 
(i) for each $\ci{X_a}{X_b}{X_U,x_W}$ that is true in $\I_{x_W}$, 
(ii) for each factor $\phi_i(X_{C_i}) \in \{\phi_i(X_{C_i})\}_i$, it holds
that (iii) either $a \notin C_i$ or $b \notin C_i$.

To conclude then, we argue that conclusions (i), (ii) and (iii) of the Auxiliary Lemma are equivalent to the
requirements (a), (b), and (c) of the factorization. Clearly, conclusions (i) and (iii)
matches requirement (a) and (c) of the factorization. We now show the equivalence of (ii) with (b). 
A factor $\phi_i(X_{C_i})$ of the conditional $p(X \setminus X_W \mid x_W)$ is equivalent
to a factor $\phi_i(X_{C_i}, x_W)$ over the joint $p(X)$, which is composed of features 
$f_{C_i \cup W}$ whose values over $X_W$ matches $x_W$, i.e., 
$\project{f_{C_i \cup W}}{W} = x_W$. 
\end{proof}
\end{theorem}

The theorem requires that each possible reduced dependency model of $\I_c$
be graph-isomorph. What is the implication of this requirement? 
By definition of graph-isomorphism, this implies that for each possible context
$x_W$, the reduced dependency model $\I_{x_W}$ can be encoded as an undirected graph
over the sub-domain $X \setminus X_W$. This provides us a mean to construct 
$\I_c$ graphically, i.e., constructing an undirected graph for each possible
sub-domain and assignment of its complement. In practice, this may be done
by experts that provide a list of CSIs that hold in the domain, or running a
structure learning algorithm over each context. This may sound overly complex,
as there are cleary an exponential number of such contexts. No doubt future
works can explore this aspect, finding alternatives for simplifying this complexity
on different special cases.

\section{Conclusions} \label{sec:conclusions}

We have presented a theoretical method for factorizing a Markov random field
according to the CSIs present in a distribution, that is formally guaranteed to be correct.
This is presented by the context-specific Hammersley-Clifford theorem,
as a generalization to CSIs of the Hammersley-Clifford theorem that applies for conditional independences.
According with our theoretical result, we believe that it is worth 
guiding our future work in implementing algorithms for learning from data the structure of MRFs for each possible context,
and then factorizing the distribution by using the learned structures.
Intuitively, it seems likely to achieve improvements in time,
space and sample complexities, in comparison with other approaches
that only uses conditional independences encoded by the graph.







\end{document}